\documentclass[sigconf]{acmart}
\AtBeginDocument{%
  \providecommand\BibTeX{{%
    \normalfont B\kern-0.5em{\scshape i\kern-0.25em b}\kern-0.8em\TeX}}}

\acmConference[IARPA BETTER]{Intelligence Advanced Research Projects Activity -- 
Better Extraction from Text Towards Enhanced Retrieval}{March 23,
  2023}{University of Maryland, College Park}
\begin{document}

\title{An Interactive Query Generation Assistant using LLM-based Prompt Modification and User Feedback}
\author{Kaustubh D. Dhole}
\email{kdhole@emory.edu}
\affiliation{%
    \institution{Department of Computer Science}
  \institution{Emory University}
  \city{Atlanta}
  \country{USA}
}
\author{Ramraj Chandradevan}
\email{rchan31@emory.edu}
\affiliation{%
    \institution{Department of Computer Science}
  \institution{Emory University}
  \city{Atlanta}
  \country{USA}
}
\author{Eugene Agichtein}
\email{yagicht@emory.edu}
\affiliation{%
    \institution{Department of Computer Science}
  \institution{Emory University}
  \city{Atlanta}
  \country{USA}
}
\renewcommand{\shortauthors}{Dhole, Chandradevan \& Agichtein}
\begin{abstract}
While search is the predominant method of accessing information, formulating effective queries remains a challenging task, especially for situations where the users are not familiar with a domain, or searching for documents in other languages, or looking for complex information such as events, which are not easily expressible as queries. Providing example documents or passages of interest, might be easier for a user, however, such query-by-example scenarios are prone to concept drift, and are highly sensitive to the query generation method. This demo illustrates complementary approaches of using LLMs interactively, assisting and enabling the user to provide edits and feedback at all stages of the query formulation process. The proposed \textbf{Query Generation Assistant} is a novel search interface which supports automatic and interactive query generation over a mono-linguial or multi-lingual document collection. Specifically, the proposed assistive interface enables the users to refine the queries generated by different LLMs, to provide feedback on the retrieved documents or passages, and is able to incorporate the users' feedback as prompts to generate more effective queries. The proposed interface is a valuable experimental tool for exploring fine-tuning and prompting of LLMs for query generation to qualitatively evaluate the effectiveness of retrieval and ranking models, and for conducting Human-in-the-Loop (HITL) experiments for complex search tasks where users struggle to formulate queries without such assistance.
\end{abstract}
\maketitle

\vspace{5pt}
\section{Introduction}
Retrieving information is critically important from documents in multiple languages as the Internet increasingly provides access to documents across thousands of languages and  domains. Creating effective search queries can be a daunting task for users. First, users may be unfamiliar with the language of the information they need to obtain or may be completely unaware of it, making it hard to craft specific queries. Second, most people are not familiar with the vocabulary and jargon used in other areas or fields, which can further impair their ability to formulate good search queries.  Furthermore, users may be unfamiliar with the corpus, or collection of documents being searched, making it challenging to know what information to look for and how to phrase the information need.
We propose a solution to this challenge by employing ``query-by-example'' - allowing users to explore document collections better by letting them specify an example document (rather than an explicit query) of what they are searching for. Although considerable advancements have been made in the domain of query-by-example~\cite{qbe1,qbe2}, especially neural and transformer based, there is a clear lack of interfacing tools for performing qualitative analysis making it an area ripe for exploration.

As query-by-example (QBE) and multilingual information retrieval (MLIR) introduce new tasks to the traditional information retrieval community, new research questions and challenges arise, such as how to provide effective search results in different languages and how to assist users in generating effective queries. Traditional methods of qualitative analysis can be time-consuming, as researchers need to manually generate queries and analyze search results, making it even harder to iterate. Hence, having an interfacing tool that can automatically generate queries and display the search results together can be invaluable for researchers and practitioners alike.

On the other hand, success in few-shot prompting~\cite{srivastava2022imitation, brown2020language,pretrainprompt} has led large language models to play a key role in reducing the information burden on users by especially assisting them for writing tasks namely essay writing, summarisation, transcript and dialog generation, etc. This success has also been transferred to tasks related to query generation~\cite{jeong-etal-2021-unsupervised, nogueira2019document}. While large language model applications are prevalent and numerous studies have been conducted for search interfaces~\cite{s1,s2,s3,s4}, there has been little impetus to combine search interfaces with large language model based query generation. 

In this paper, we demonstrate \textbf{Query Generation Assistant}, a search interface that supports automatic and interactive query generation for monolingual or multi-lingual interactive search. The novel contributions of the proposed interface include: 
\begin{enumerate}
    \item The interface provides a simple document search interface that displays documents in their original language along with their translations, making it simple for researchers to navigate and analyse search results.
    \item The tool also supports diverse query generation, allowing users to explore search results more comprehensively.
    \item More importantly, it combines search with a prompting-based query generation interface which permits users to refine their queries and prompts with retrieval information.
\end{enumerate}
We believe our interface could work as an effective starting template for performing qualitative analysis over other search related experiments and datasets as well as serve as a tool to incorporate retrieval feedback and Human-In-The-Loop (HITL) studies. Even though our system was built initially for the BETTER search tasks (described in the next subsection,) 
our interface is generic in nature and would be transferable to other datasets and indices. We share the python code for the below described interface as well as the video demonstration here\footnote{\url{https://github.com/emory-irlab/better-search}}.

We first briefly describe the BETTER task and dataset in Section 1. We then explicate the three main features of Query Generation Assistant in section 3.

\section{Dataset}
Our system and interface was designed to investigate interactive query generation especially for Query-By-Example (QBE) settings. 
The BETTER search datasets\footnote{\url{https://ir.nist.gov/better/}}~\cite{mckinnon-rubino-2022-iarpa,bettercrosslang} were used for demonstration. The BETTER dataset is a collection of natural language processing resources developed by IARPA’s BETTER program\footnote{\url{https://www.iarpa.gov/research-programs/better}} to assist their intelligence analysts to process and analyze huge amounts of unstructured, multilingual information efficiently and effectively and serves as an example application for multi-lingual QBE and document retrieval for event monitoring (event retrieval). This collection also contains ancillary information like event span annotations from text across many languages and topics. Particularly, the BETTER program seeks search systems to perform accurate retrieval of Arabic, Persian, Chinese, Korean, and Russian documents on being queried with example English documents. 

\section{Query Generation Assistant}
The Query Generation Assistant User Interface is made up of 3 subsystems. Each of them are described below. The interface is built using HuggingFace's Gradio platform. Gradio~\cite{wolf-etal-2020-transformers, abid2019gradio} is an open-source Python package to quickly create easy-to-use, customizable UI components for machine learning models.

\subsection{Manual Search}
\begin{figure*}
  \includegraphics[width=\textwidth]{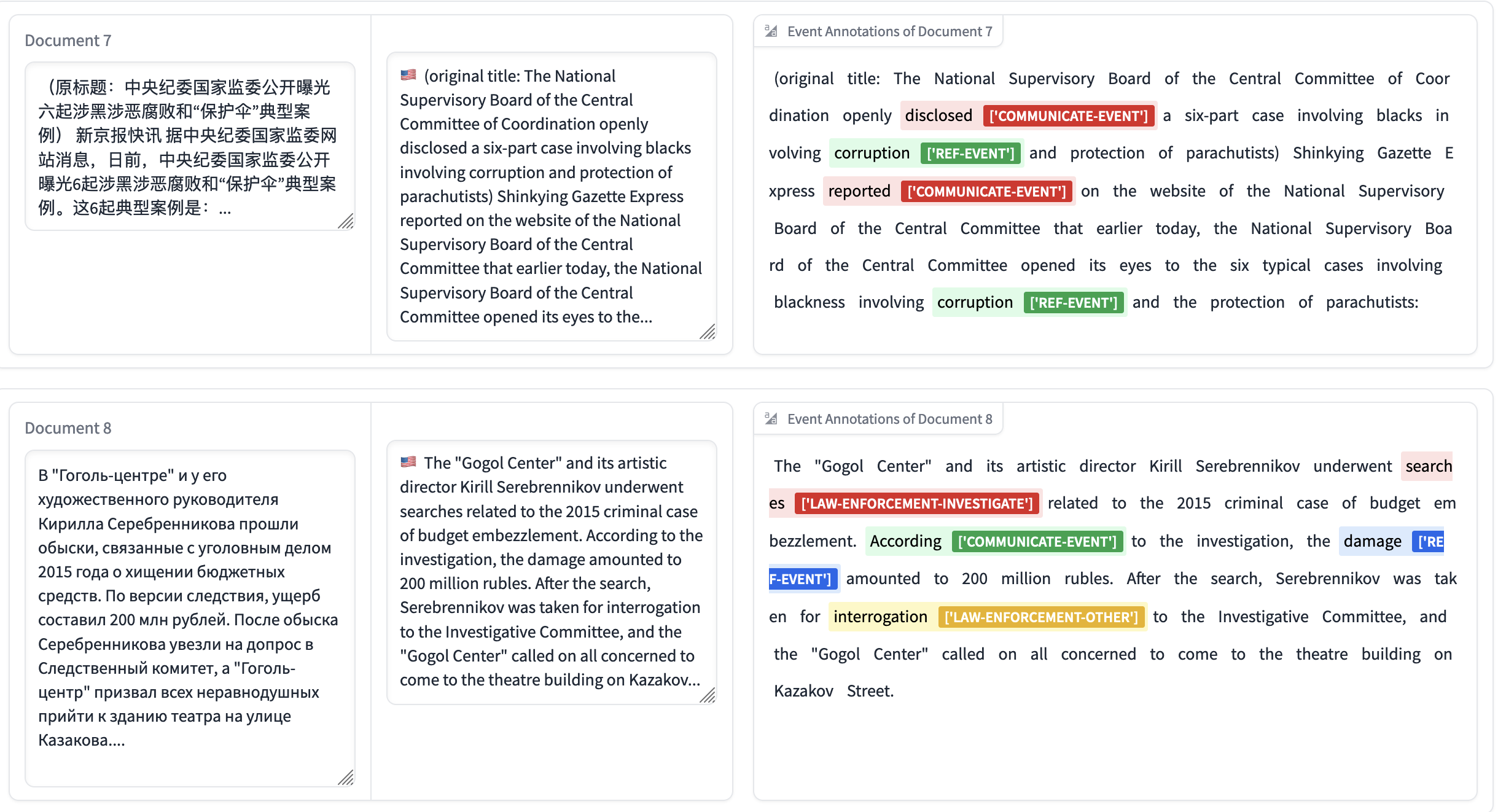}
  \caption{Manual Search: The list of returned documents displayed alongwith their translations and event annotations.}
  \label{fig_manual_search}
\end{figure*}
This tab (shown in Figure~\ref{fig_manual_search}) is the simplest interface which permits users to write search queries by themselves and returns the top-k relevant documents, their English translations, and the highlighted events for each document. The search is conducted on each index built using their corresponding language's documents, using SOTA cross-lingual dense retrieval model, ColBERT-x~\cite{colbert-x}. A document rank list is returned for each language. These rank lists are eventually combined and reranked using reciprocal rank fusion. All the documents are translated offline using Google Translate\footnote{\url{https://translate.google.com/}} for faster look-up during query time. To highlight the event annotations, such as event triggers and argument entities, on the displayed documents, the collection is parsed to
a SOTA event annotator (span-finder~\cite{span-finder}) offline and then looked-up during the query time. 

\subsection{Auto Query Generator}
The BETTER task seeks to benchmark systems to be able to look for documents in specified target languages which are similar to a user’s example document. We attempt to do this via generating an intermediate query from the example document and performing retrieval over the same. However, the effectiveness of the generated queries is crucial to retrieve relevant documents, while also ensuring query interpretability.

Inspired by the recent success of pre-trained generation models, we fine-tune a T5~\cite{raffel2020exploring} model on (document, query) pairs. To evaluate the performance of our approach, we compare the original T5 model with a docT5query~\cite{nogueira2019document} model, which has already been fine-tuned on the MSMarco~\cite{nguyen2016ms} dataset. Our results indicate that the docT5query model outperforms the original T5 model, and thus we utilize it for our demonstration. 

\begin{figure*}
  \includegraphics[width=\textwidth]{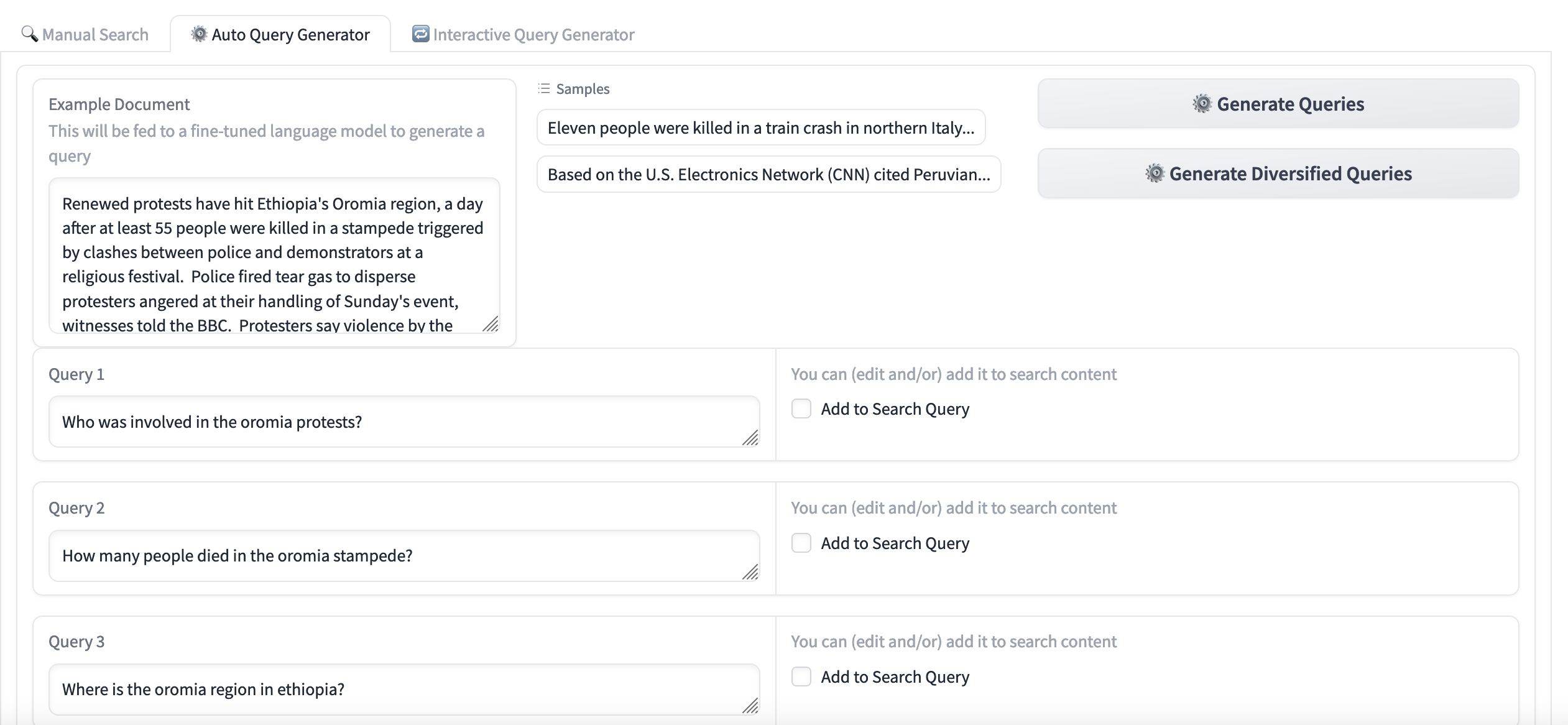}
  \caption{Auto Query Generator: Given an example document, queries are generated. Queries can be further refined by the user and searched across the documents.}
  \label{fig:teaser1}
\end{figure*}

The complete interface is shown in Figure~\ref{fig:teaser1}.

\subsection{Interactive Query Generation}
Large language models in recent years have shown excellent strides in multi-task learning and few-shot learning. With just a handful of examples, large language models have shown impressive generative capabilities, albeit with risks of hallucinations. Prompting has been an effective and seemingly natural way to interact with such models. 

While few-shot prompting has been a powerful approach to teach models new tasks, such models have generally shown different outputs on varying various prompting parameters. For example, varying the types of examples in the prompt, changing the order of the examples and number of examples vastly influence the generations. We use this feature to our advantage for query generation by letting people edit their prompts either directly or through user relevance feedback so as to improve subsequent query generations and corresponding retrieval. 

\begin{figure*}  \includegraphics[width=\textwidth]{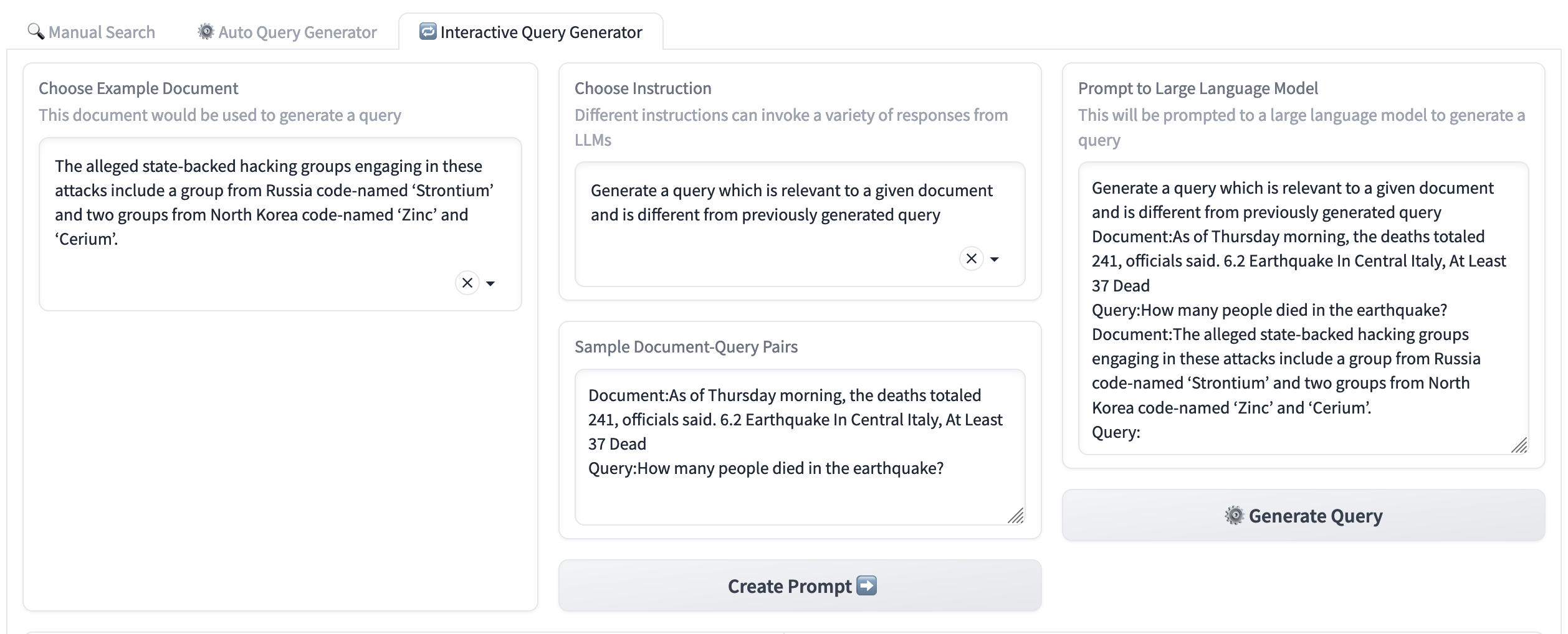}
  \caption{Prompt Based Interactive Query Generator}
  \label{fig:teaser2}
\end{figure*}

We choose FlanT5~\cite{chung2022scaling} as it has been fine-tuned already on a large amount of tasks making it arguably convenient~\cite{aribandi2022ext} for learning on new tasks. The interface permits prompting FlanT5 by default with two editable (document, query) pairs alongwith an instruction. We present users with options of multiple instructions and their choice of document to generate query from. The interface is shown in Figure~\ref{fig:teaser2}. 

Each of the generated queries can at once or together be used to retrieve documents. On retrieval, each of the retrieved document is provided with a checkbox to permit it to be directly added to the prompt alongwith its original query. This is intended to incorporate user search feedback directly into the prompt to make the prompt more consistent with the users' requests. In terms of few-shot prompting, when models are prompted to generate responses based on a limited set of examples, the quality of the generations depends on the quality and relevance of the examples provided as models are known to be also less robust towards prompt perturbations\cite{zhao2021calibrate,dhole2023nl}.

\section{Conclusion}
The primary objective of Query Generation Assistant is to assist researchers with an ability to qualitatively monitor cross-lingual retrieval and provide assistance to generate and refine queries. Researchers and practitioners can quickly and easily perform qualitative analysis with the tool's search interface and query generation features, allowing them to evaluate search systems more thoroughly. The prompting-based search interface also provides an avenue to perform human-in-the-loop (HITL) studies. Apart from qualitative studies, we believe Query Generation Assistant could be used as an effective starting template to perform more sophisticated information retrieval experiments as well as serve as a tool to incorporate retrieval feedback and conduct Human-In-The-Loop studies.

\section{Acknowledgements}
This work was supported in part by IARPA BETTER (\#2019-19051600005). The views and conclusions contained in this work are those of the authors and should not be interpreted as necessarily representing the official policies, either expressed or implied, or endorsements of ODNI, IARPA, or the U.S. Government. The U.S. Government is authorized to reproduce and distribute reprints for governmental purposes notwithstanding any copyright annotation therein.

\bibliographystyle{ACM-Reference-Format}
\bibliography{sample-base, software}

\end{document}